%% file: main.tex
\documentclass[conference]{IEEEtran}
\IEEEoverridecommandlockouts
\usepackage{cite}
\usepackage{amsmath,amssymb,amsfonts}
\usepackage{algorithmic}
\usepackage{graphicx}
\usepackage{textcomp}
\usepackage{multirow}
\usepackage{comment}
\usepackage{url}
\usepackage{wrapfig}
\usepackage{subcaption}
\usepackage{diagbox}
\usepackage{xcolor}
\usepackage{hyperref}
\begin{document}

\title{A Converting Autoencoder Toward Low-latency and Energy-efficient DNN Inference at the Edge\\
\thanks{}
}

\author{\IEEEauthorblockN{Hasanul Mahmud, Peng Kang, Kevin Desai, Palden Lama and Sushil K. Prasad}
\IEEEauthorblockA{Department of Computer Science \\ University of Texas at San Antonio\\
San Antonio, Texas\\
Email: \{hasanul.mahmud, peng.kang, kevin.desai, palden.lama, sushil.prasad\}@utsa.edu}}

\maketitle

\begin{abstract}

Reducing inference time and energy usage while maintaining prediction accuracy has become a significant concern for deep neural networks (DNN) inference on resource-constrained edge devices. To address this problem, we propose a novel approach based on ``converting" autoencoder and lightweight DNNs. This improves upon recent work such as early-exiting framework and DNN partitioning. Early-exiting frameworks spend different amounts of computation power for different input data depending upon their complexity. However, they can be inefficient in real-world scenarios that deal with many hard image samples. On the other hand, DNN partitioning algorithms that utilize the computation power of both the cloud and edge devices can be affected by network delays and intermittent connections between the cloud and the edge. We present CBNet, a low-latency and energy-efficient DNN inference framework tailored for edge devices. It utilizes 
a ``converting" autoencoder to efficiently transform hard images into easy ones, which are subsequently processed by a lightweight DNN for inference. To the best of our knowledge, such autoencoder has not been proposed earlier. Our experimental results using three popular image-classification datasets on a Raspberry Pi 4, a Google Cloud instance, and an instance with Nvidia Tesla K80 GPU show that CBNet achieves up to 4.8$\times$ speedup in inference latency and 79\% reduction in energy usage compared to competing techniques while maintaining similar or higher accuracy.

\end{abstract}


\input{Introduction}

\input{RelatedWork}

\input{Design}

\input{Evaluation}

\input{Conclusion}

%\section*{Acknowledgment}
%The research is supported by NSF CNS 1911012 grant and UTSA.

\bibliographystyle{abbrv}
\bibliography{main}

\input{Appendix}
\end{document}

%% file: introduction.tex
\section{Introduction}

Over the past decade, the field of machine learning has achieved tremendous success in classifying images and videos\cite{article}. These impressive advances in machine learning have relied heavily on using more powerful hardware (e.g. GPUs) and more complex software such as deep neural networks (DNNs). While DNN inference can achieve high accuracy, this comes at the cost of increasing execution time (latency) and energy consumption. These issues are becoming more prominent for deep learning at the edge ~\cite{chen2019deep,li2018network}. Executing deep learning on edge devices provides low-latency access to data sources and also alleviates data privacy concerns. However, it is challenging to accommodate the high resource requirements of DNN inference on resource-constrained edge computing resources\cite{EdgeAI}.

\begin{comment}
\begin{figure}[tp]
  \begin{subfigure}[t]{0.23\textwidth}
      \centering
		\includegraphics[width=\textwidth]{figure/hard_easy.png}
    	\caption{MNIST dataset.}
  \end{subfigure}
  \hfill
  \begin{subfigure}[t]{0.23\textwidth}
      \centering
		\includegraphics[width=\textwidth]{figure/LocalServer1/easy_hard2.pdf}
    	\caption{Fashion-MNIST dataset.}
  \end{subfigure}
  \caption{ Easy image compared with hard image in image-classification datasets. For each figure, the left image is easier to classify than the right image. }
  \label{fig:easy-hard-image}
\end{figure}
\end{comment}

\begin{figure}[tp]
  \begin{subfigure}[t]{0.23\textwidth}
      \centering
		\includegraphics[width=\textwidth]{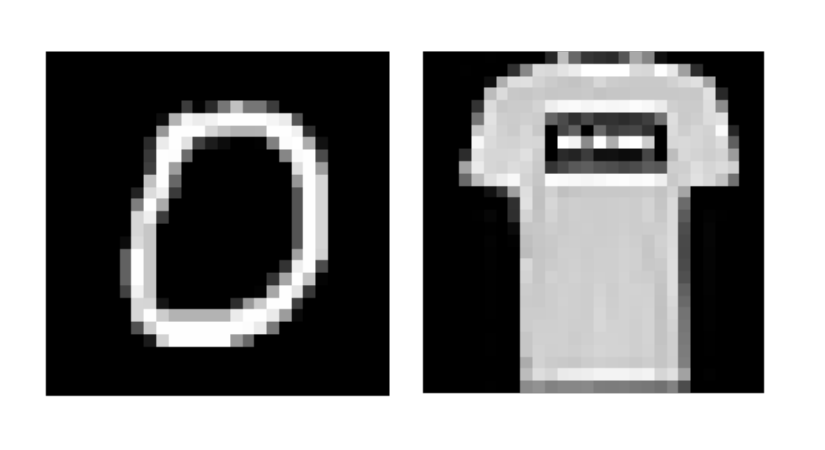}
    	\caption{Easy images.}
  \end{subfigure}
  \hfill
  \begin{subfigure}[t]{0.23\textwidth}
      \centering
		\includegraphics[width=\textwidth]{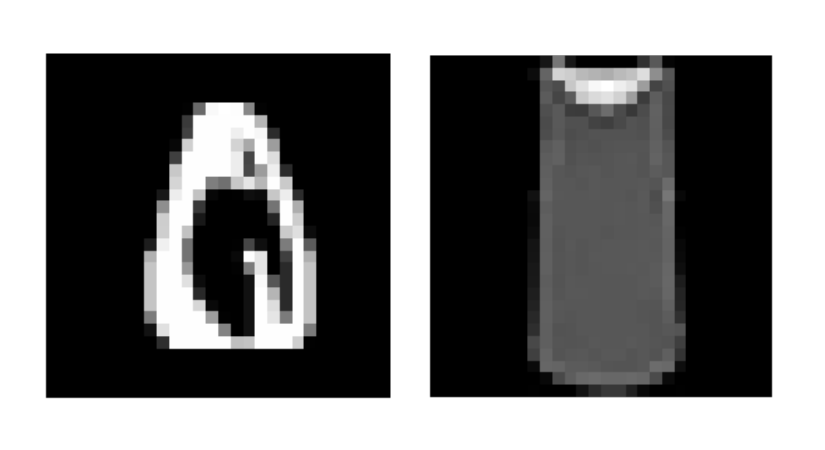}
    	\caption{Hard images.}
  \end{subfigure}
  \caption{Easy images compared with hard images in image-classification datasets, MNIST and Fashion-MNIST.}
  \label{fig:easy-hard-image}
\end{figure}

Recent works propose DNN model partitioning to offload some layers of the model and its computation from the edge devices to the cloud backend~\cite{Kang2017NeurosurgeonCI,li2018edge}. However, such approaches can be affected by network delays and intermittent connections between cloud and edge devices. Another line of research has focused on early-exiting DNN frameworks where results from computing the initial
layers are used to provide approximate classification results~\cite{panda2016conditional,teerapittayanon2017branchynet}. These leverage the idea that some data samples can be classified with much less work than others. For example, Fig.~\ref{fig:easy-hard-image} illustrates how input images in the image-classification datasets MNIST and Fashion-MNIST can differ in complexity. The images of a handwritten digit 0 and a T-shirt, shown in Fig.~\ref{fig:easy-hard-image} (a), are easy to classify by computing only the initial layers of a DNN model. Whereas, the corresponding images shown in Fig.~\ref{fig:easy-hard-image} (b) may require computation of the entire DNN for accurate classification. Although early-exit DNN is a promising approach, it can still be inefficient in real-world scenarios that deal with a large number of hard inputs in the dataset. Hard inputs can range from low-resolution or blurry images to complex images that are dissimilar to other images belonging to the same class. 

In this paper, we present CBNet, a low-latency and energy-efficient framework for DNN inference on edge devices. We refer to edge devices as compute units ranging from reasonably provisioned edge servers equipped with a GPU, to smartphones with mobile processors, to barebone Raspberry Pi devices. The core component of CBNet is a \emph{converting autoencoder} that is designed to transform hard images into easy images. It learns efficient data encoding of hard input images and recovers easy images of the same class. The images transformed by the autoencoder are then fed to a lightweight DNN for inference. The inference latency of CBNet is the sum of the execution time spent in the autoencoder and the lightweight DNN classifier. Starting with an early-exit DNN such as BranchyNet~\cite{teerapittayanon2017branchynet}, we developed a general approach to training the \emph{converting autoencoder} as well as for obtaining the corresponding lightweight DNN. As a proof of concept, we evaluated our framework for three popular image-classification datasets including MNIST\cite{deng2012mnist}, Fashion-MNIST (FMNIST)\cite{xiao2017fashionmnist}, and Kuzushiji-MNIST (KMNIST)\cite{clan2018deep}.

%For this, we trained the BranchyNet framework and collected the exit points for corresponding input. For BranchyNet , all inputs finish the computation taking lower two exits(i.e. exit 0 and 1). While training via Autoencoder, we assumed that inputs take zero exits as easier inputs and one as more complex inputs. We showed that our results are  better than the  BranchyNet results in terms of inferencing time and energy consumption. 

Our key contributions in this paper are as follows:
\begin{itemize}
    %\item We explored to use several classical Machine Learning models, including SVM, MLP, etc, and the newest technique, locality-sensitive hashing, to solve the problem how to accurately classify the complex images and simple images. These methods didn't work well when solving this problem.

    %\item We designed a framework, CB-LeNet, to use simple autoencoder technique to encode/decode complex images to the simple images and use a less complex Neural network to inference input images. On the other hand, CB-LeNet need  less training time compared with BranchyNet since CB-LeNet use more simple Neural Network. \\

    \item We design and train a novel \emph{converting autoencoder} model to encode a hard image into an efficient representation that can be decoded into an easy image belonging to the same class.
    \item We design and implement CBNet, an efficient framework for DNN inference on resource-constrained edge devices. CBNet utilizes the \emph{converting autoencoder}, which coupled with a lightweight DNN classifier facilitates fast and energy-efficient inference without compromising accuracy. %This demonstrates the potential of our novel approach based on   inputs and truncating DNNs. 
    
    %\item The CB-LeNet framework and the methodology presented in this paper are general enough to be adapted to various DNN architectures and datasets. As a proof of concept, we use the DNN architecture of the LeNet~\cite{Lecun1998Lenet} model as a baseline.
    
    %\item We evaluated our framework at different platforms, including real edge devices(Raspberry Pi 4) running in the Chameleon Chi-Edge platform and edge server mimicked by a virtual machine running in the Google cloud. As a result, our framework reduces the energy consumption by 40\% and 72\% for Raspberry Pi 4 and Google cloud server, respectively compared to the BranchyNet-LeNet model. \\
    
    \item Experiments conducted on a Raspberry Pi 4 and a Google Cloud instance with and without Nvidia Tesla K80 GPU, show that CBNet 
    achieves up to 3.82$\times$, 3.78$\times$ and 4.8$\times$ speedup in inference latency compared to BranchyNet~\cite{teerapittayanon2017branchynet}, AdaDeep\cite{Adadeep} and SubFlow\cite{subflow} respectively, while maintaining similar or higher accuracy. 
    %\item Our scalability analysis shows that the rate of increase in inference latency with the increasing size of the dataset in the case of CBNet is significantly lower than that of competing approaches.
    \item CBNet reduces the total energy usage associated with DNN inference on a Raspberry Pi 4 device, a Google Cloud instance, and an instance with GPU by up to 73\%, 71\%, and 79\%, respectively, compared to  BranchyNet. While compared to the baseline LeNet~\cite{Lecun1998Lenet} model, it reduces the energy usage on these devices by up to 85\%, 81\%, and 81\%, respectively. 
    
    %\item We investigated the power consumption model for different scenarios/settings. 

    %Apart from our proposed framework CB-LeNet, we also explored several classical Machine Learning models, including Support Vector Machine(SVM), Multi-Layer Perceptron(MLP), Decision Tree(DT), etc., and a hashing technique called Locality-sensitive hashing(LSH) to classify input images into two categories: Easy and Complex. Unfortunately, these experiments did not bring fruitful results.
    %\item Verify our experiment at three different scenarios, including real edge devices, Raspberry Pi 4.
    %\item Verify our experiment at three different platforms %including real edge devices, Raspberry Pi 4.
\end{itemize}

The remainder of the paper is organized as follows. Section \ref{Background} describes the background and related work. Section \ref{Implementation} describes the key design and implementation details. Section \ref{Evaluation} details the testbed setup and experimental results. Section \ref{Conclusion} discusses the conclusion and future work.

%% file: RelatedWork.tex
\section{Background and Related work}\label{Background}

In recent years, several studies have focused on reducing the computational burden of DNN inference on resource-constrained devices. We discuss some major approaches that have been proposed to address this research problem.

%Deep learning network has experienced most success among all machine learning algorithms in solving complex problems.On the other hand, A deep learning network may be accurate but it is resource-intensive, especially for edge/end devices. Nowadays, lots of data are being produced by some intelligent applications like alexa,siri,cortana etc. used by end users. In this section, we will discuss different approaches used for training and inferencing these datas.

%\subsection{Cloud Only Approach} 
%One of the most common approaches used by cloud providers for intelligent applications is to perform all DNN processing in the cloud. A significant overhead of this approach is in sending data over the wireless network\cite{Bakshi}. Extensive amount of data are sent from end device to cloud server for necessary computation, and then inferencing results sent back to the user/end-devices.

\subsection{Autoencoders}
An autoencoder is an artificial neural network that learns efficient encodings of unlabeled data. It consists of an Encoder, which learns how to efficiently encode data into a reduced representation, and a Decoder, which learns how to reconstruct the data back to a representation that is as close to the original input as possible~\cite{vincent2010stacked}. There are several variations of autoencoders. The denoising autoencoders are trained to recover original input from intentionally perturbed or noisy input\cite{Goodfellow-et-al-2016}, with the aim to learn a more robust representation of input data. 
%On the other hand, 
A variational autoencoder is a generative model that can produce different variations of existing data samples~\cite{vae}. A transforming autoencoder was introduced in ~\cite{transforming_autoencoder} that learns encoded representations of not only the object's identity in an image but also its pose, such as position, orientation, scale, and lighting. By applying different transformations to the input image, the model learns to adjust its parameters accordingly while preserving the object's identity. Autoencoders are also widely used for dimensionality reduction, denoising, data augmentation, and anomaly detection. Our synergistic system approach sets us apart, developing a novel \emph{converting autoencoder} trained for hard-to-easy image transformation and feeding the converted images to the early-exit framework, reducing latency and energy usage.

\subsection{Early-exit DNN}
%commented
%Early-exiting framework exploits the idea that forcing the samples to take early exits that can learn at early stages of the network can save inference time by some extent. In this section, we will briefly discuss about a few recent research progresses in this area. Panda et al.~\cite{panda2016conditional} proposed Conditional Deep Learning (CDL) where the convolutional layer features are used to identify the variability in the hardness or complexity of input instances and conditionally activate the deeper layers of the network. In CDL, auxiliary linear classifiers are iteratively added to each convolutional layer, starting with the first layer and the output is observed to decide whether to exited early. The proposed methodology enables the network to dynamically adjust the computational effort depending upon the complexity of the input data while maintaining competitive classification accuracy. Teerapittayanon et al.~\cite{teerapittayanon2017branchynet} proposed BranchyNet as an early-exit DNN that allows for more general branch network structures with additional layers at each exit point. BranchyNet allows prediction results for test samples to exit the network early via these branches if the samples can be inferred with high
%confidence. Unlike CDL~\cite{panda2016conditional}, it jointly trains the branches with the original network to  significantly improve the performance of the overall architecture.%

An early-exiting framework exploits the idea that some data samples can be classified accurately even if they pass through only the initial few layers of the neural network. 
In this section, we will briefly discuss about a few recent research efforts in this area. Panda et al.~\cite{panda2016conditional} proposed Conditional Deep Learning (CDL) where the convolutional layer features are used to identify the variability in the hardness or complexity of input instances and conditionally activate the deeper layers of the network. In CDL, auxiliary linear classifiers are iteratively added to each convolutional layer, starting with the first layer and the output is observed to decide whether to exit early. The proposed methodology enables the network to dynamically adjust the computational effort depending upon the complexity of the input data while maintaining competitive classification accuracy. Teerapittayanon et al.~\cite{teerapittayanon2017branchynet} proposed an improved version of early-exiting framework named BranchyNet where they added branches in the neural network at various exit points. BranchyNet allows prediction results for test samples to exit the network early via these branches if the samples can be inferred with high confidence. Unlike CDL~\cite{panda2016conditional}, it jointly trains the branches with the original network to significantly improve the performance of the overall architecture. %Fig.~\ref{fig:branchynet} illustrates a simple BranchyNet architecture where an early-exit branch is added to the main branch of LeNet~\cite{Lecun1998Lenet}, a baseline neural network. 
At each exit point, BranchyNet uses the entropy of a classification result (e.g., by softmax) as a measure of confidence in the prediction. If the entropy of a test sample is below a learned threshold value, meaning that the classifier is confident in the prediction, the sample exits the network with the prediction result at this exit point, and is not
processed by the higher network layers. If the entropy value is above the threshold, then the classifier at this exit point is deemed not confident, and the sample continues to the next
exit point in the network.

%For each sample input, it will check the confidence value whether it can be inferred correctly at \textquotedblleft Exit 1\textquotedblright{} else it will take later \textquotedblleft Exit 2\textquotedblright.

\begin{comment}

\begin{figure}[tp]
    \centering
    \includegraphics[width=9cm]{figure/branchynet.png}
    \caption{B-LeNet: BranchyNet Architecture based on LeNet}
    \label{fig:branchynet}
\end{figure}
\end{comment}

\begin{figure*}[ht]
      \centering
      \includegraphics[width=0.8\textwidth]{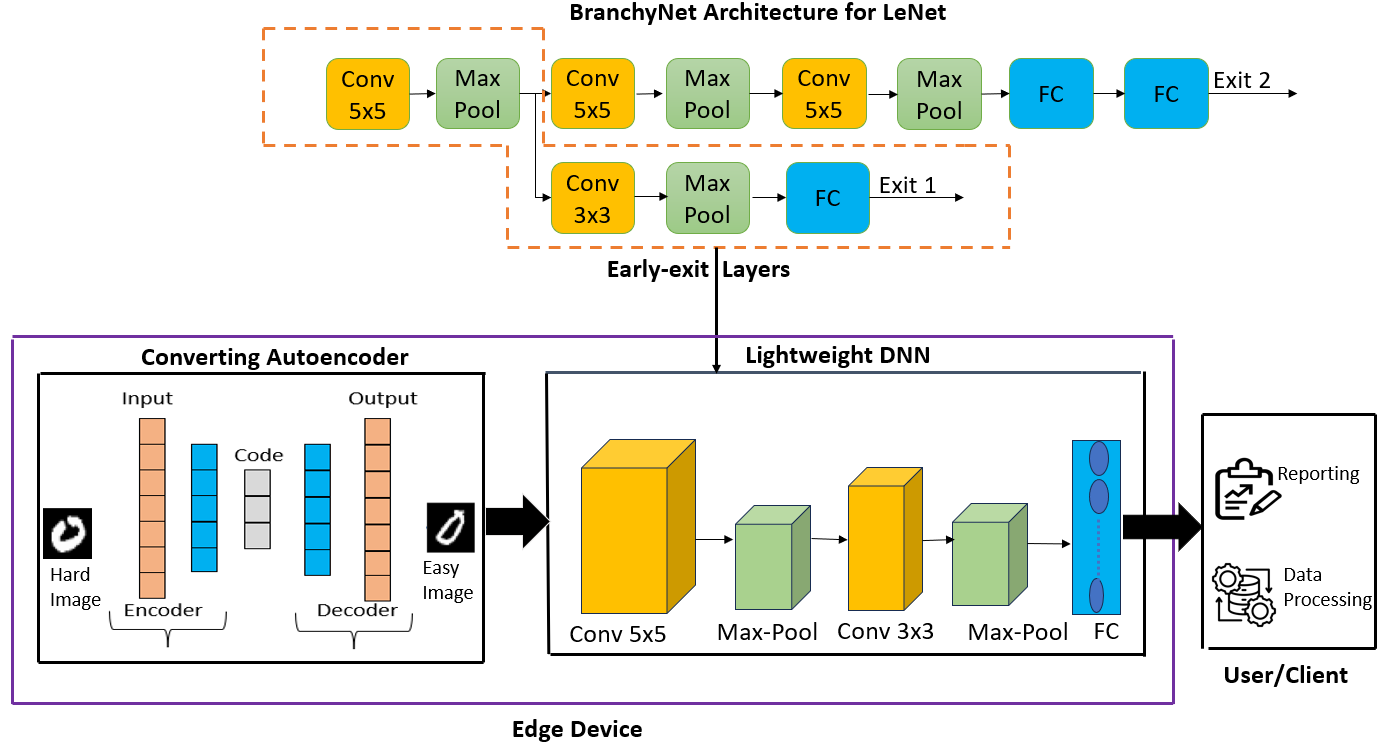}
  \caption{System Overview:- A ``converting" autoencoder transforms hard images into easy images of the same class. The resulting images are fed to a lightweight model extracted from an early-exit DNN for inference.}
  \label{fig:Architecture}
\end{figure*}

Jayakodi et al.~\cite{Jayakodi_2018} proposed an optimization algorithm to automatically configure Coarse-to-Fine Networks (C2F) Nets for a specified trade-off between accuracy and energy consumption. Phuong et al.~\cite{Phuong2019DistillationBasedTF} proposed a new training procedure for multi-exit architectures based on the principle of knowledge distillation. Their training method encouraged early exits to mimic later, more accurate exits, by matching their output probabilities. In~\cite{Park-DAC2020}, the authors proposed a Branch-pruned Conditional Neural Network (BPNet) and it's methodology that determines how many or where the auxiliary classifiers or branches should be added in a systematic fashion to tradeoff inference time and accuracy. Le et al.~\cite{Yang2020ResolutionAN} proposed a Resolution Adaptive Network (RANet), which is based on the intuition that low-resolution representations are sufficient for classifying “easy” inputs containing large objects with prototypical features, while only some “hard” samples need spatially detailed information. A major limitation of existing early-exit DNNs is that they can be inefficient when a significant portion of the dataset consists of hard inputs. We address this issue with a novel approach based on \emph{converting autoencoder}.
%Fig.\ref{fig:confidence_branchy} depicts a LeNet architecture adding one lower branch as suggested in BranchyNet. At each exit points, the input will check whether it is confident enough to predict accurately. These early exit branches allow samples which can be accurately classified in early stages of the network to exit at that stage.
\subsection{DNN Partitioning}
%Processing only in the Cloud require significant amounts of data to be sent to the cloud over the wireless network and puts tremendous computational pressure on the data center. Nowadays, as the computational power and resources in mobile devices has become more powerful, researchers worked to see whether this cloud-only processing is desirable to move forward, and what are the implications of pushing some or all of this compute to the mobile devices on edge.

%In most of the AI applications, user/input datas are sent to cloud from the end/edge devices for training and inferencing and the output is sent back to the user device.In recent times running machine learning algorithm like DNN on resouce constrained edge device has gained significant attention.But, running DNN on this end devices has an overhead of high computation and energy consumption. E.% 

DNN partitioning is an alternative technique that reduces the inference time by offloading some layers of the neural network model and its computation from low-end mobile edge devices to an edge server or the cloud. 
%In this section, we described a brief details of some related research works involving DNN partitioning. 
Li et al.~\cite{li2018edge} proposed a DNN co-inference framework that adaptively partitions DNN computation between mobile devices and the edge server based on the available bandwidth, thus taking advantage of the processing power of the edge server while reducing data transfer delay.
In another work\cite{Kang2017NeurosurgeonCI}, a lightweight scheduler called Neurosurgeon was designed to automatically partition DNN computation between mobile devices and data centers at the granularity of neural network layers. Zhou et al.~\cite{9546452} developed a dynamic-programming-based search algorithm to decide the optimal partition and parallelization for a DNN model. Unlike DNN partitioning-based approaches, our work focuses on improving the latency and energy-efficiency of DNN inference on a single edge device. Our approach avoids the issues associated with network latency, intermittent connection and the overheads of DNN re-partitioning in a dynamic environment.  

%Even though DNN partitioning has some advantages over cloud-only approach in terms of latency and energy-savings, it needs to take into account an additional algorithm for selecting DNN partitioning point at any given time. This produces an additional bottleneck with regards to execution time.

\begin{comment}
\graphicspath{{figure/Easy and Hard Lenet}} 
  \begin{figure}[t]
    \centering
    \includegraphics[width=4cm]{figure/Branchynet_confidence.PNG}
    \caption{BranchyNet framework. If Inference model has enough confidence, it will take the corresponding early exits.}
    \label{fig:confidence_branchy}
    
  \end{figure}
  
 \begin{figure*}[ht]
      \centering
    \includegraphics[width=0.85\textwidth]{figure/architecture.pdf}
  \caption{EncodeNet architecture. }
  \label{fig:Architecture}
\end{figure*}
\end{comment}

\subsection{DNN Compression}
DNN compression techniques can reduce the computation and storage requirements of a DNN model. In \cite{10.1145/3297858.3304011}, the authors proposed two tools: SONIC and TAILS. SONIC significantly reduces the cost of guaranteeing correct intermittent execution for loop-heavy code like DNN inference, whereas TAILS exploits SIMD (Single Instruction Multiple Data) hardware available in some edge devices to improve energy efficiency. Han et al.~\cite{Han2016DeepCC} proposed a three-stage pipeline: pruning, trained quantization, and Huffman coding, to compress the DNN layers. The network pruning technique is widely used in DNN compression to make neural networks lightweight and energy-efficient\cite{yao2017deepiot,aghasi2017nettrim}. Energy aware pruning was introduced in \cite{Energywareprune} by removing the layers based on the energy consumption of neural network layers. AdaDeep\cite{Adadeep} was proposed as a usage-driven, automated DNN compression framework for systematically exploring the desired trade-off between
performance and resource constraints. AdaDeep automatically selects the most suitable combination of compression techniques and the corresponding compression hyperparameters for a given DNN. SubFlow\cite{subflow} induced subgraph approach to fulfill the execution of a DNN task within a time constraint. This approach aims to maintain comparable performance to executing the entire network by running a subset of the DNN during runtime.
These techniques are complementary to our work and can be used in conjunction with our approach. Furthermore, evaluation results show that our approach outperforms AdaDeep\cite{Adadeep} and Subflow\cite{subflow} in terms of inference latency and accuracy.

%% file: Design.tex
\section{Design and Implementation} \label{Implementation}
In this section, we present the design and implementation of CBNet.
Fig.~\ref{fig:Architecture} shows the architecture of the CBNet framework. The key components include a \emph{converting autoencoder} that transforms hard images into easy images of the same class, which are fed to a lightweight DNN classifier for inference. %We now discuss a general approach that we developed to train a \emph{converting autoencoder} and obtain the corresponding lightweight DNN. 

%We increased the threshold value(i.e., the suitable threshold value to acquire the best accuracy in the BranchyNet framework) to enforce more input data into the early exit. However, if we use the threshold value used by BranchyNet for best accuracy, a significant portion of data takes the later exits. Consequently, the performance, including accuracy and inference time, does not improve. Since the original threshold value is based on raw input data, the early layer exit has lower confidence to inference input data. However, once data is passed into Autoencoder, most complex data will be transformed into relatively easier data and taken out from the early exit.  

%I don't get this point. Writing should have some flow.Please add this if you think this are okay.
%
%We also checked whether using the translated data to train BranchyNet can improve accuracy. The results were not impressive as we expected.Therefore, CB-LeNet can be directly running beyond the existed model. Training of BranchyNet framework buys a significant amount of time. For this reason,  training and testing methodology of our experiment are same as the BranchyNet model.

%We only use data translated by Autoencoder to inference and do not use it to train any model. 

%Then, early exit should have a higher confidence to inference the data.

\begin{figure}[t]
    \centering
    \includegraphics[width=0.36\textwidth]{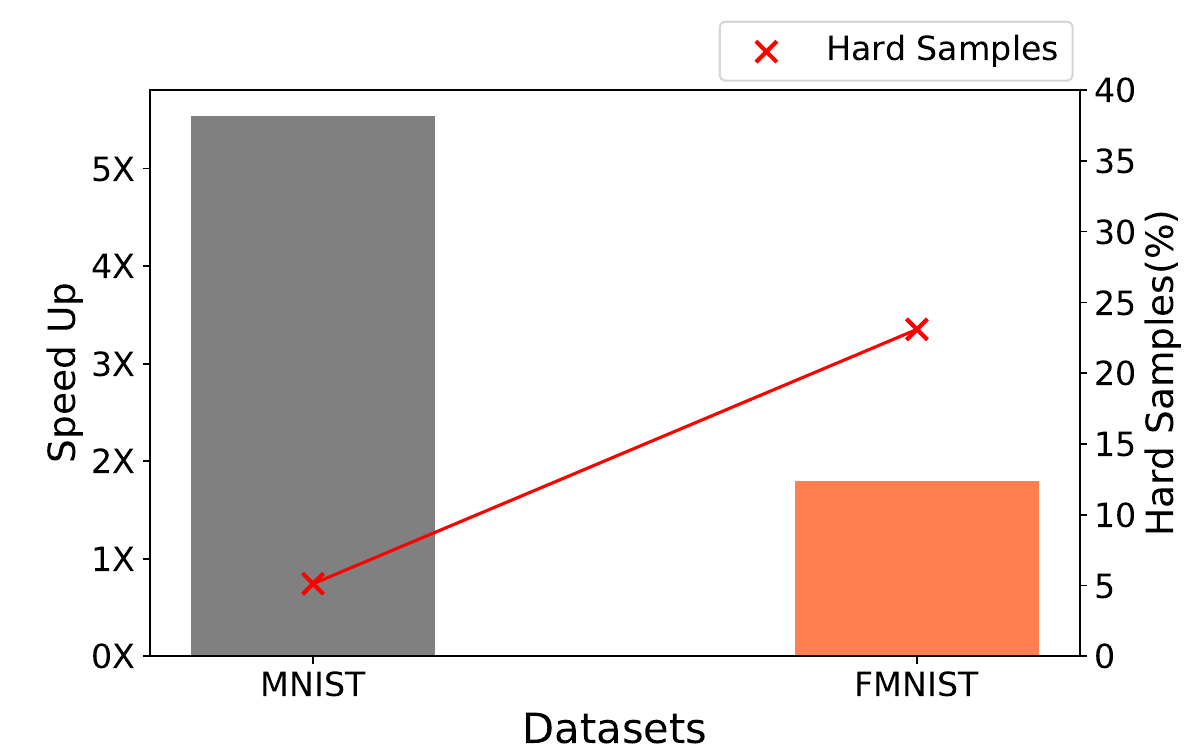}
    \caption{BranchyNet's inference speedup over LeNet diminishes when a significant fraction of the dataset contains hard images. Inference time was measured on a Raspberry Pi 4.}
\label{fig:Reltive_speedup_branchyNettoLeNet}
\end{figure}

\subsection{Converting Autoencoder}
\subsubsection{Motivation for hard-to-easy image transformation}
We design and train a \emph{converting autoencoder} model to encode a hard image into an efficient representation that can be decoded into an easy image belonging to the same class. The need for such image transformation is motivated by the inefficiency of existing early-exit DNNs in dealing with hard images. As a case study, we evaluated the speedup in inference latency achieved by BranchyNet\cite{teerapittayanon2017branchynet} over the baseline LeNet model for various datasets on a Raspberry Pi 4. As shown in Figure~\ref{fig:Reltive_speedup_branchyNettoLeNet}, BranchyNet's inference speedup drops from 5.5$\times$ in the case of MNIST~\cite{deng2012mnist} dataset to 1.7$\times$ in the case of FMNIST~\cite{xiao2017fashionmnist} dataset. This is because 23\% of data samples in FMNIST are hard images that are processed by deeper layers of BranchyNet. Whereas the hard images constitute only 5\% of the MNIST dataset. We aim to address this issue with our \emph{converting autoencoder} that can efficiently achieve hard-to-easy image transformation to facilitate fast inference without compromising accuracy.

\subsubsection{Training the Converting Autoencoder using BranchyNet}

\begin{figure}[t]
\centering
\includegraphics[width=0.50\textwidth]{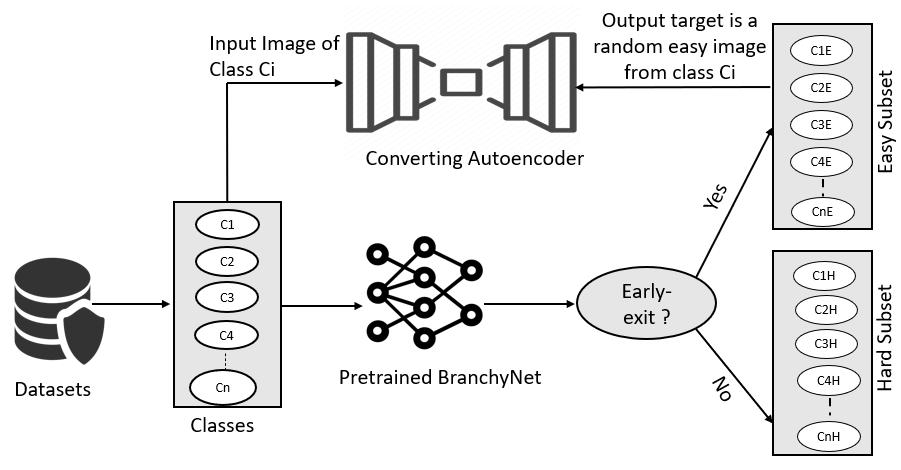}
\caption{Training ``converting" autoencoder for image transformation.}
\label{fig:training_process}
\end{figure}

The training of our \emph{converting autoencoder} requires images that are labeled as easy or hard. However, publicly available image classification datasets do not come with such labels. Therefore, we used the BranchyNet model~\cite{teerapittayanon2017branchynet} to prepare the training data with appropriate labels. As shown in Fig.~\ref{fig:training_process}, we passed images from the training dataset through a pre-trained BranchyNet model for inference. We labeled the images that exited the network early as easy images and labeled the rest as hard images. All images (both hard and easy) were then passed through the \emph{converting autoencoder} as training input. For each image as input, an easy image that belongs to the same class was randomly chosen as the target output. Finally, the reconstruction loss was calculated as the mean squared error between the model output and the target output. %that had the lowest entropy of classification result as the target output. Having lower entropy means that BranchyNet is more confident in the prediction. 
%We observe that the reconstruction loss can be minimized when the easiest image from each class is used as the target output for that class. For each class, the easiest image is the one for which BranchyNet has the highest confidence in prediction. BranchyNet measures the confidence in prediction by calculating the entropy of the classification result.   

%We have shown in Table \ref{tab:accuracy}, the percentages of exits show the percentage of easy data and hard data, and accuracy of the each model using four different datasets. When training via autoencoder, we picked out the simplest data, which has the lowest value of entropy loss function, to make all input data to fit the simplest data. We designed our autoencoder model in such a way that it takes complex images to transform those into more simpler images.
%We trained autoencoder with the easy portion of training data and then convert the test original datasets into easier test datasets using trained model from autoencoder. In the end, these converted images are then passed into the Branchynet framework for inferencing. Our framework CB-LeNet exhibited promising results in terms of inference time and energy-consumption compared to the Branchynet without compromising any accuracy.   

\subsubsection{Hyperparameter Tuning}

\begin{comment}

\begin{tt}
\begin{table}[tp]
    \centering
    \caption{Hyperparameter search space.}

    \begin{tabular}{|c|c|}
    \hline
          Hyperparameter   & Search Space   \\ 
        \hline
            activation & ['relu', 'linear', 'sigmoid', 'tanh']   \\
            Optimizer & ['SGD', 'Adam'] \\
            units of first layer & [784, 512, 384]  \\
            units of second layer & [384, 256, 128]  \\
            units of third layer & [128, 64, 32]  \\
            activity regularizer  & [L1(10e-6), L1(10e-8)] \\
            %\\gridSearchCV didn't include batch size and epochs in the hypyerparameter tunning. We need manually adjust epochs and batch size.
            % talos is a tool for keras, which can include batch size and epochs. The software compatible 
            % problem need to consider. 
            % after hpyerparameter tunning, we manually use different batch size to run gridSearch again and find best choices. 
            batch size  & [128, 256, 512]  \\
            %epochs  & [10, 20, 50, 100] \\
    \hline
    \end{tabular}
    \label{tab:HyperparameterTuning}
\end{table}
\end{tt}

\end{comment}

One of our key design goals is to minimize the computational overhead of the \emph{converting autoencoder} while limiting the reconstruction loss. This is to ensure that the execution time and energy usage overhead associated with the autoencoder is less than the improvement in the latency and energy usage associated with DNN inference. For this purpose, 
we designed a lightweight model with only three hidden layers. 
%Furthermore, we used the Grid Search\footnote{\url{https://scikit-learn.org/stable/modules/generated/sklearn.model_selection.GridSearchCV.html}} method to tune the hyperparameters of the model. The search space includes the activation function, optimizer, the number of units in each layer, activity regularizer, and batch size as shown in Table \ref{tab:HyperparameterTuning}. %We initialized the epochs and batch size as 20 and 256 before the search started. 
Table \ref{tab:autoencode-architecture} shows the autoencoder architecture optimized for various image-classification datasets. Each autoencoder uses the Adam optimizer~\cite{kingma2014adam} to update the neural network weights during the training process. Furthermore, it adds penalties to the reconstruction loss function in proportion to the magnitude of the activations in the output of the Encoder layer. This process, known as activity regularization, improves the robustness and generalization of the model. As the activity regularizer, we used L1 penalty with a coefficient of 10e-8.

\subsection{A lightweight DNN Classifier for Inference}
%\subsection{Truncated BranchyNet}

We use a lightweight DNN to process the images transformed by the \emph{converting autoencoder}. In this work, the DNN is obtained by truncating the early-exit branch of BranchyNet shown in Figure~\ref{fig:Architecture}. The lightweight DNN consists of 2 convolutional layers and 1 fully connected layer. For non-BranchyNet DNNs with layers 1 through $N$, a truncated network (layer 1 through $k<N$) appended with a suitable output layer can be employed as a lightweight DNN. Experimentally, a reasonable number of layers K can be found iteratively starting with $K=1$, guided by the resulting number of hard and easy images in a dataset, and the corresponding complexity of the autoencoder.

%% file: Evaluation.tex
\section{Evaluation} \label{Evaluation}
\subsection{Experimental TestBed} \label{TestBed}

\begin{footnotesize}
\begin{tt}
\begin{table*}[tp]
    \centering
    \caption{Converting autoencoder architecture trained on various datasets.}

    \begin{tabular}{|c|c|c|c|c|c|c|}
    \hline
         \multirow{2}{*}{\backslashbox[8em]{layer}{hyperparameter}} &  \multicolumn{2}{c|}{MNIST}  & \multicolumn{2}{c|}{FMNIST}  & \multicolumn{2}{c|}{KMNIST} \\
        \cline{2-7}   &  size of feature map &  activation function &  size of feature map &  activation function &  size of feature map &  activation function  \\ 
        \hline
            Input   & 784 & - & 784 & - & 784 & - \\
            FullyConnected1    
                    & 784 & relu &  512& relu &    512 & relu \\
       
            FullyConnected2
                    & 384 & relu   & 256  & relu  &   384 & linear \\
        
            FullyConnected3
                    & 32  & linear & 128 & linear &  32 & linear  \\
            FullyConnected4 & 784 & Softmax & 784 & Softmax & 784 & Softmax \\

    \hline
    \end{tabular}
    \label{tab:autoencode-architecture}
\end{table*}
\end{tt}
\end{footnotesize}

We repeated our experiments three times across two testbeds; Google Cloud\footnote{\url{https://cloud.google.com/compute/docs/instances}}, and Chameleon CHI@Edge platform\cite{keahey2020lessons}. We used a general-purpose Google Cloud instance (GCI), N1 machine series, running Ubuntu 16.04, equipped with 2 vCPUs and 8 GB RAM and a similar cloud instance equipped with a Nvidia Tesla K80 GPU.
We used a Raspberry Pi 4 device, equipped with 4 CPU (ARM V8) cores and 8GB RAM, hosted on the Chameleon CHI@Edge platform. To evaluate our framework on Raspberry Pi, we built a Docker container image with arm64v8/ubuntu:18.04 as the base image, Keras library\footnote{\url{https://blog.keras.io/building-autoencoders-in-keras.html}}, scikit-image-arm64v8\footnote{\url{https://github.com/tutu-kang/scikit-image-arm64v8}} library and BranchyNet library. The scikit-image-arm64v8 is an image processing library that supports arm64v8 architecture. The original BranchyNet library has a dependency on the scikit-image library that only supports x86 processors. Therefore, we modified the BranchyNet library to use scikit-image-arm64v8 instead. We launched the Docker Container (Engine Version 18.06.2-ce) to run our experiments on the Raspberry Pi 4.

\subsection{Evaluation Models and Datasets} 
\subsubsection{Evaluation models}\label{Sec:EvaluationModel}

We compared CBNet with various competing models including baseline LeNet\cite{Lecun1998Lenet},   BranchyNet-LeNet (BranchyNet)\cite{teerapittayanon2017branchynet}, AdaDeep\cite{Adadeep} and SubFlow\cite{subflow}. 
LeNet~\cite{Lecun1998Lenet} is a convolutional neural network for image classification. We used a specific type of BranchyNet called BranchyNet-LeNet which is based on the LeNet~\cite{Lecun1998Lenet}  architecture. BranchyNet consists of three convolutional layers and two fully-connected layers in the main network. It has one early-exit branch consisting of one convolutional layer and one fully-connected layer after the first convolutional layer of the main network. In our experiments, the BranchyNet confidence threshold is set to 0.05 for the MNIST datasets, 0.5 for FMNIST and 0.025 for KMNIST. These thresholds were tuned to achieve the maximum performance for BranchyNet. AdaDeep\cite{Adadeep} and SubFlow\cite{subflow} are two recent frameworks for DNN compression. 

%Truncated BranchyNet is essentially similar to Ours without the \emph{converting autoencoder}.

%This is equivalent to a BranchyNet where the main network is cut-off and only the early-exit branch is active. Truncated BranchyNet is essentially similar to Ours without the \emph{converting autoencoder}.  

%We compared the energy efficiency and performance of Ours with LeNet\cite{Lecun1998Lenet}, Branchy-LeNet (BranchyNet)\cite{teerapittayanon2017branchynet}, and Truncated BranchyNet. Table~\ref{tab:LeNet_architecture} shows the LeNet architecture. It consists of three convolutional layers and two fully-connected layers in the main network. Table~\ref{tab:BLeNet_architecture} shows the architecture of Truncated BranchyNet which consists of 2 convolutional layers and 1 fully connected layer. This is equivalent to a BranchyNet, shown in Figure~\ref{fig:Architecture}, where the main network is cut-off and only the early-exit branch (Exit 0) is active. Truncated BranchyNet is essentially similar to Ours without the \emph{converting autoencoder}.  
%BranchyNet uses LeNet as the main network and Truncated BranchyNet as a branch, which happened after the first common convolutional layer. Ours uses four fully-connected layers(the first three layers for encoding and the last layer for decoding) as autoencoder as shown in Table \ref{tab:autoencode-architecture} and the remaining layers have the same architecture with Truncated BranchyNet. 

\subsubsection{Datasets} We evaluated CBNet and other competing models on three image-classification datasets: MNIST~\cite{deng2012mnist}, FMNIST~\cite{xiao2017fashionmnist}, and KMNIST~\cite{clan2018deep}. The MNIST dataset is a large dataset of handwritten digits that is commonly used for training various image processing system. It includes a training set of 60,000 examples, and a test set of 10,000 examples. The digits have been size-normalized and centered in a fixed-size image. Fashion-MNIST~(FMNIST) is a dataset of Zalando's article images--consisting of a training set of 60,000 examples and a test set of 10,000 examples. Each example is a 28x28 grayscale image, associated with a label of 10 types of clothing, such as shoes t-shirts, dresses, and more. Kuzushiji-MNIST~(KMNIST), which focuses on $Kuzushiji$~(cursive Japanese), is a drop-in replacement for the MNIST dataset, It contains 70,000 28x28 grayscale images spanning 10 classes(one from each column of hiragana), and is balanced like original MNIST dataset (6k/1k train/test for each class).

%The MNIST, FMNIST, and KMNIST datasets consist of a training set of 60000 examples and a test set of 10000 examples each. Whereas QMNIST consists of 60000 training and 60000 test examples. We trained and tested each model separately for each dataset. Fig. \ref{fig:Ours_training} shows the training process of Ours's \emph{converting autoencoder}. We observe that the training and validation loss decreases and converges to a steady value as the training progresses through multiple epochs. 
%relationship of epoch and loss function, including the training losses and validation losses, for autoencoder. We use this point of intersection(the training loss and validation loss converged after several epochs) as the training epochs for MNIST and FMNIST. We chose the lowest point of a parabola(the training loss decreased continually as validation loss forms a parabola) as the training epochs for KMNIST. The training epochs for MNIST, FMNIST, and KMNIST are 100, 35, and 6. 

\subsection{Modeling Energy and Power Consumption}
To evaluate the energy efficiency of CBNet, we used separate power consumption models for the GCI without GPU and the Raspberry Pi 4 device hosted on Chameleon CHI@Edge platform. These platforms do not provide the facility for direct power measurement. In the case of the GCI with Nvidia GPU K80, we directly measured the average power usage of the GPU by using the \emph{nvidia-smi} tool. We measured the energy usage ($E$), in Joules, as a product of the average power ($P$) consumption, in Watts, and inference latency ($\Delta t$), in seconds. Unlike competing models, the inference latency of CBNet is measured by adding up the execution time of the 
converting autoencoder and the DNN inference time. %In the case of other models competing with CBNet, effective inference latency is equal to the DNN inference time. 
Power consumption can be influenced by many factors including CPU usage, Memory access, I/O access, temperature, etc. For the sake of simplicity, we only focus on CPU power modeling since DNN inference is mostly CPU-bound. %When measuring the inference time, we didn't calculate the data transfer time. Therefore, we use the inference time as the elapsed time. 

\subsubsection{GCI CPU power modeling} 
We used the CPU power model shown in Equation~\ref{eq:cpuPower} to estimate the power consumption of the GCI~\cite{Isci2003Micro, Hsu2011ISPASS, LIN2018FGCS}. Power consumption, $P_{GCI}$, is calculated as: 
\begin{equation}
\label{eq:cpuPower}
\begin{aligned}
P_{GCI} = (n/N) \times ( P_{idle} +  \left ( P_{peak} - P_{idle}  \right ) \times u_{v}^{\beta } )
\end{aligned}
\end{equation}
where $n=2$ denotes the number of virtual CPUs (vCPUs) allocated to the GCI. $N=18$ denotes the numbers of CPU cores available on the physical server where the cloud instance is hosted. $P_{idle}$ and $ P_{peak}$ respectively denote the idle power and the peak power consumption of the physical server. $u_v$ represents vCPU runtime utilization that can be monitored via guest-OS APIs. We used the tool, \emph{psutil}, to measure the CPU utilization. $\beta$ is an exponent of the power function model. Similar to ~\cite{Hsu2011ISPASS}, we used a $\beta$ value of 0.75. 
The physical server that hosts our GCI, N1 machine series, is equipped with Intel Xeon E5-2699 V3 (Haswell) processor. Therefore, we consider $P_{idle}$ and $ P_{peak}$ to be 40W and 180W respectively. These values were reported by Wang et al. in~\cite{wang2015iwomp}.

\begin{figure}[ht]
    \centering
    \includegraphics[width=0.4\textwidth]{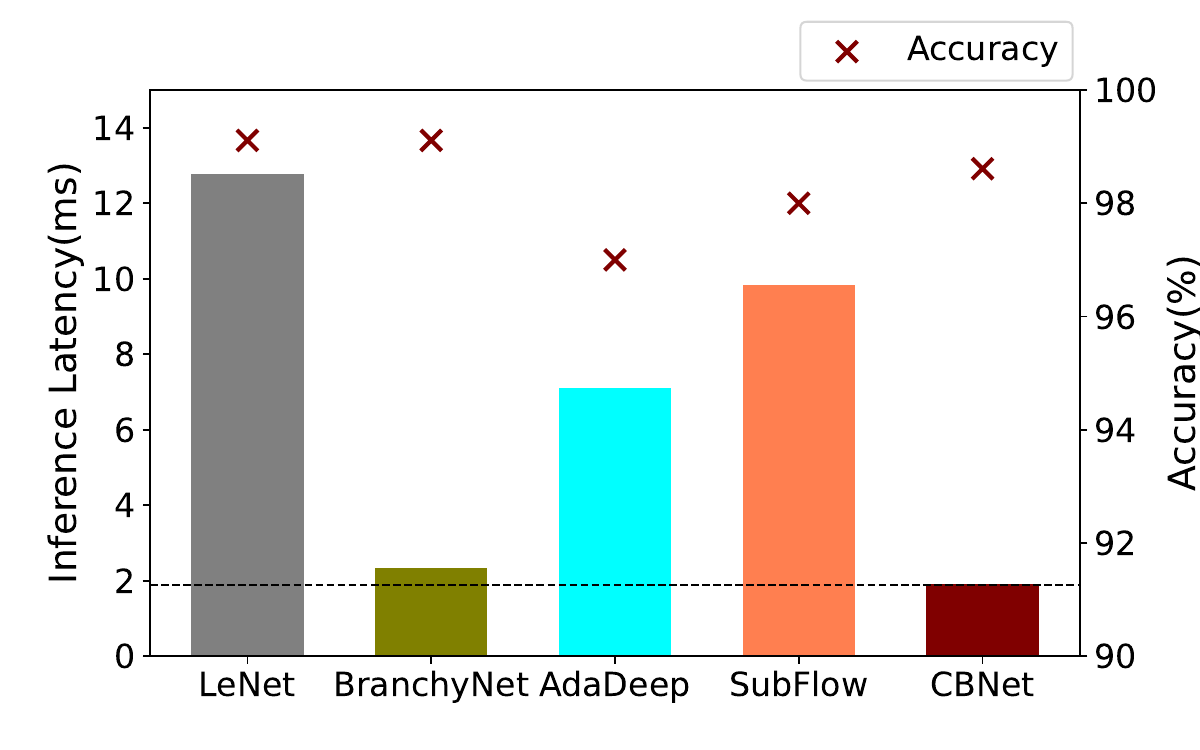}
    \caption{Comparing CBNet with baseline, BranchyNet, AdaDeep\cite{Adadeep}, and Subflow\cite{subflow} in terms of inference latency and accuracy for MNIST dataset on Raspberry Pi 4.}
    \label{fig:Inference_latency_state_of_the_art}
\end{figure}

%combined table
\begin{tt}
\begin{table*}[ht]
    
    \centering
    \caption{Comparing CBNet with baseline and BranchyNet in terms of inference latency, energy efficiency, and accuracy.}

    \begin{tabular}{c|c|c|c|c|c|c|c|c}
    \hline
    %Dataset & Model  & \multicolumn{3}{C | }{} Raspberry Pi 4 (s) & GCI w/o GPU (s) & GCI with GPU (s) & Accuracy (\%)\\
    \centering {Dataset} & Model  & \multicolumn{3}{c |} {Latency per image (Milliseconds)} & \multicolumn{3}{c |} {Energy Savings (\%) w.r.t LeNet} & Accuracy (\%) \\
    \cline{3-8}
    & &Raspberry Pi 4  & GCI w/o GPU  & GCI with GPU  &Raspberry Pi 4  & GCI w/o GPU  & GCI with GPU & \\
    
         \hline
        \multirow{3}{*}{MNIST}  
         & LeNet & 12.735  & 1.322& 0.266  &- &- &- & 99.11 \\ 
              & BranchyNet &2.3  &  0.395&  0.118 & 82 (\%)  &70 (\%) &55 (\%) & 99.11 \\
              & CBNet &1.877  &0.267 & 0.105 & 85 (\%)& 80 (\%)&80 (\%) &98.61 \\ 
             
        \hline 
        \multirow{3}{*}{FMNIST}  
         &LeNet  & 13.006 & 1.271& 0.261  & - & - & - & 86.53  \\ 
              & BranchyNet  & 7.248 & 0.716 & 0.225  &  45 (\%) & 43 (\%) & 14 (\%) & 87.8 \\
              & CBNet  &1.895  &0.261 & 0.104 &86 (\%) & 81 (\%)&80 (\%) & 89.59 \\ 
                        \hline
        \multirow{3}{*}{KMNIST} 
         & LeNet & 12.814 & 1.350 & 0.265 & -  & - & -& 91.4  \\ 
              & BranchyNet  & 5.682   & 0.9  &  0.24  &56 (\%) & 33 (\%)& 10 (\%) & 92.05\\
              & CBNet &1.897  & 0.26& 0.1 & 85 (\%)  & 82 (\%) & 81 (\%) & 92.5 \\ 
               
    \hline
    \end{tabular}
    \label{tab:CBNet_Performance}
\end{table*}
\end{tt}

%\subsubsection{GPU power measurement} 

%we can directly use the  The runtime power is the average power usage of all samples (one sample per second) when running the inference model. We measured the average runtime power as 79W.
%The idle power is the average power usage of 60 samples(one sample per second) when GPU is idle. The runtime power is the average power usage of all samples(one sample per second) when running the inference model. The idle power is 33W and the runtime power is 79W. 

\subsubsection{Raspberry Pi power modeling}
The power model used for the Raspberry Pi 4 device is derived from $PowerPi$~\cite{Kaup2014PowerPi}:

\begin{equation}
\begin{aligned}
P_{Pi} = P_{idle} + \left ( P_{peak} - P_{idle}  \right ) \times u^{\beta } 
\end{aligned}
\label{eq:raspberry_pi}
\end{equation}
where $P_{idle}$ is the idle power consumption of the device and $u$ is the CPU utilization of the device. $\beta$ is 1. $P_{idle}$ and $P_{peak}$ are 2.7 $W$ and $6.4W$ respectively~\footnote{\url{https://www.pidramble.com/wiki/benchmarks/power-consumption}}.

\subsection{Inference Latency and Classification Accuracy}

For each dataset, we measure the total time to process all images and compute the average inference latency per image. As shown in Table~\ref{tab:CBNet_Performance}, CBNet achieves significantly lower inference latency than LeNet and BranchyNet across all devices and datasets while maintaining similar accuracy. For Raspberry Pi 4, CBNet achieves 6.75$\times-$6.87$\times$ and 1.22$\times-$3.82$\times$ speedup in inference latency compared to LeNet and BranchyNet, respectively. For GCI without GPU, the speedup is 4.86$\times-$5.19$\times$ against LeNet and 1.47$\times-$3.46$\times$ against BranchyNet. Finally, for GCI with GPU, the speedup is  2.5$\times-$2.65$\times$ against LeNet and 1.12$\times-$2.4$\times$ against BranchyNet. 

These results indicate that the benefit of using CBNet is more pronounced in the case of resource-constrained edge devices such as Raspberry Pi. Furthermore, compared to the MNIST dataset, more complex datasets such as FMNIST and KMNIST benefit more from CBNet in terms of speedup in inference latency. This is because the FMNIST and KMNIST datasets contain more hard images than the MNIST dataset. About 94.88\% of test samples in the MNIST datasets took the early exit from BranchyNet, whereas only 76.91\% of samples in FMNIST and 63.08\% of samples in KMNIST took the early exit. Therefore, CBNet found more opportunities to reduce the inference latency of hard images for FMNIST and KMNIST. %In the case of CBNet, inference takes place on a lightweight DNN, which is equivalent to forcing all inputs to take the early exit of BranchyNet. In contrast, all test samples passed through the whole network in the case of LeNet. Despite using a lightweight DNN, CBNet maintains high classification accuracy. 

Figure ~\ref{fig:Inference_latency_state_of_the_art} illustrates that CBNet outperforms recent DNN compression techniques, AdaDeep\cite{Adadeep} and SubFlow \cite{subflow}, in terms of inference latency by 3.78$\times$ and 4.85$\times$ respectively while also achieving higher accuracy. These results were reported for the MNIST dataset on Raspberry Pi 4. Since AdaDeep and Subflow performed even worse than BranchyNet, we did not conduct further experiments on them.

%Figure~\ref{fig:Ours_time} shows the breakdown of Ours's normalized inference latency. 
Note that CBNet's inference latency is the combined processing time of the \emph{converting autoencoder} and the lightweight DNN classifier, with the former contributing up to 25\% of the total inference time.
\begin{figure*}[htb]
    \begin{subfigure}[b]{0.33\textwidth}
      \centering
    \includegraphics[width=\textwidth]{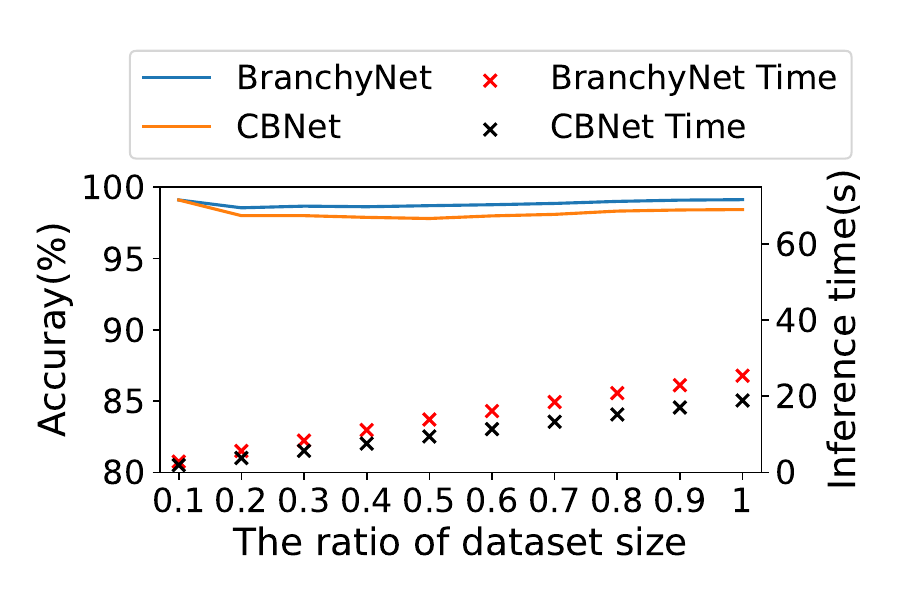}
    \caption{RaspberryPi}
    \label{fig:Raspberrpi_scale_MNIST}
    \end{subfigure}
    \begin{subfigure}[b]{0.33\textwidth}
      \centering
    \includegraphics[width=\textwidth]{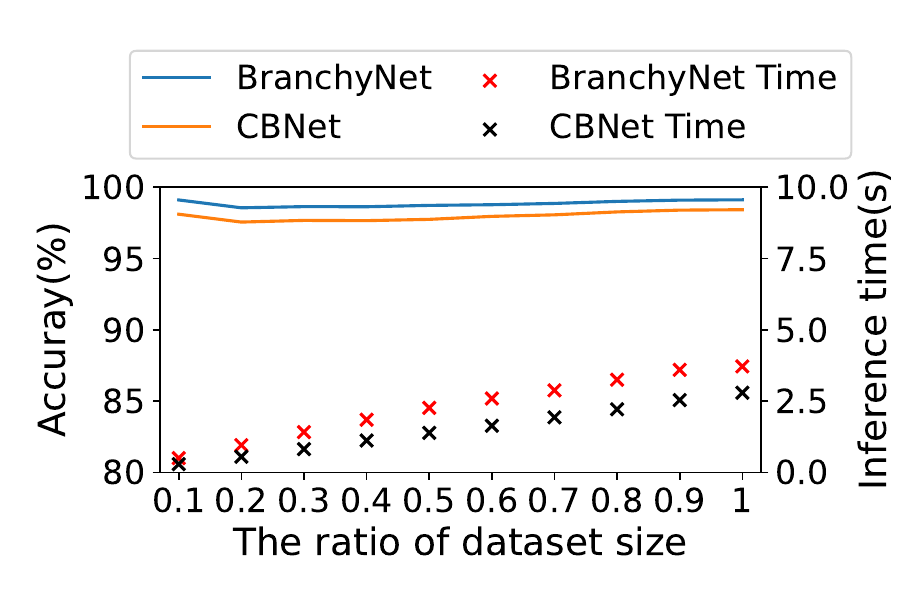}
    \caption{Google cloud instance without GPU}
    \label{fig:local_scale_mnist}
    \end{subfigure}
    \begin{subfigure}[b]{0.33\textwidth}
      \centering
    \includegraphics[width=\textwidth]{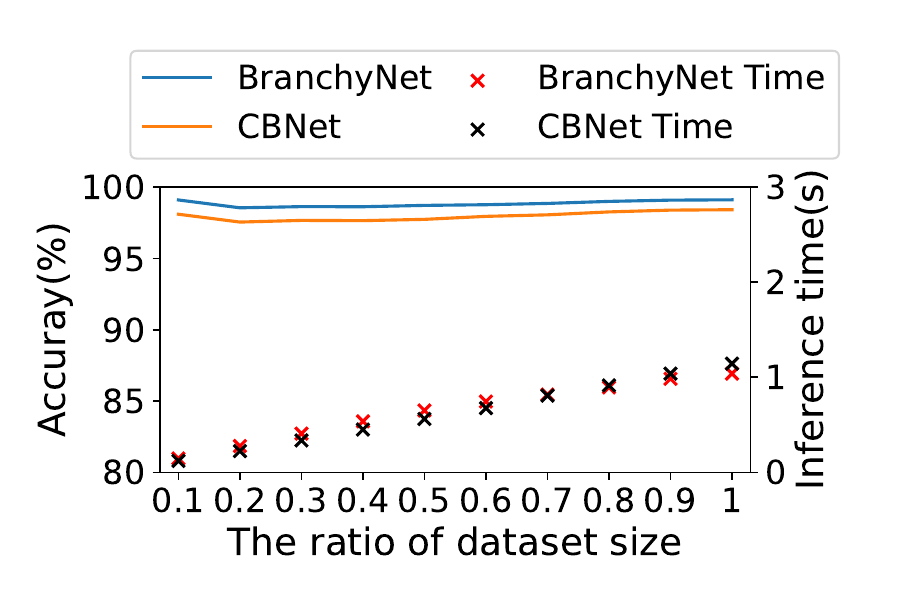}
    \caption{Google cloud instance with GPU}
    \label{fig:scale_GPU_mnist}
    \end{subfigure}
    \caption{Scalability analysis with  MNIST dataset on each hardware platform.}
    \label{fig:scale_mnist}
\end{figure*}

\begin{figure*}[htb]
    \begin{subfigure}[b]{0.33\textwidth}
      \centering
    \includegraphics[width=\textwidth]{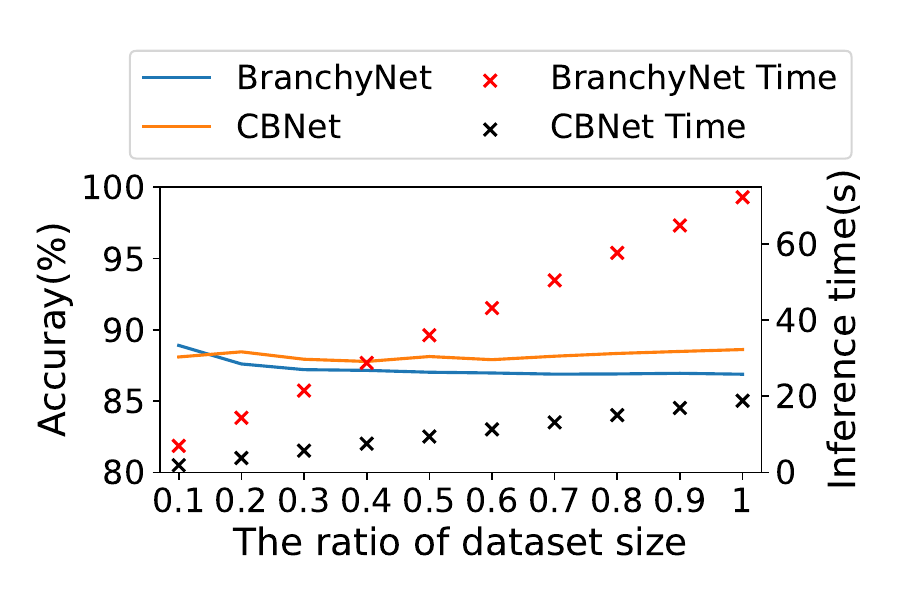}
    \caption{Raspberry Pi}
    \label{fig:Raspberrpi_scale_Fmnist}
    \end{subfigure}
    \begin{subfigure}[b]{0.33\textwidth}
      \centering
    \includegraphics[width=\textwidth]{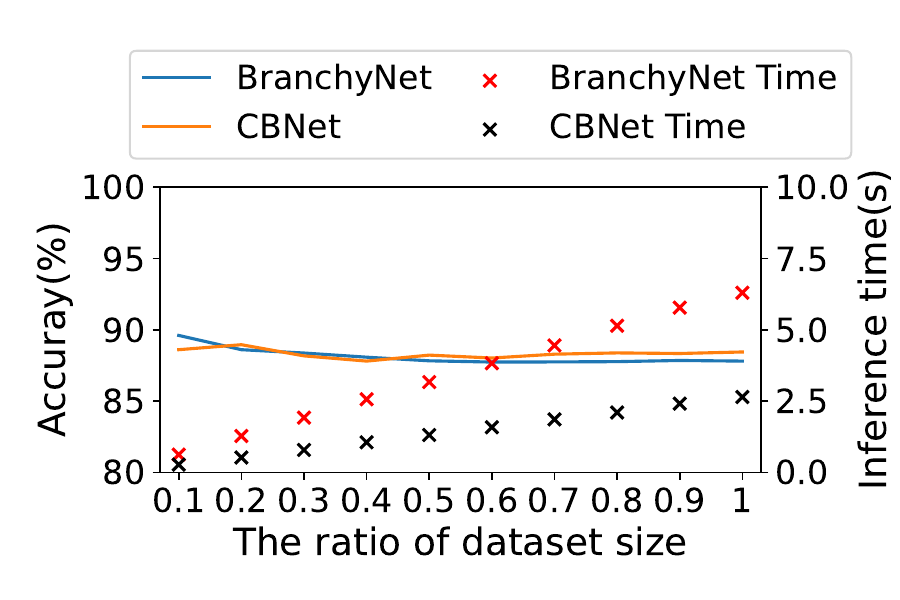}
    \caption{Google cloud instance without GPU}
    \label{fig:local_scale_fmnist}
    \end{subfigure}
    \begin{subfigure}[b]{0.33\textwidth}
      \centering
    \includegraphics[width=\textwidth]{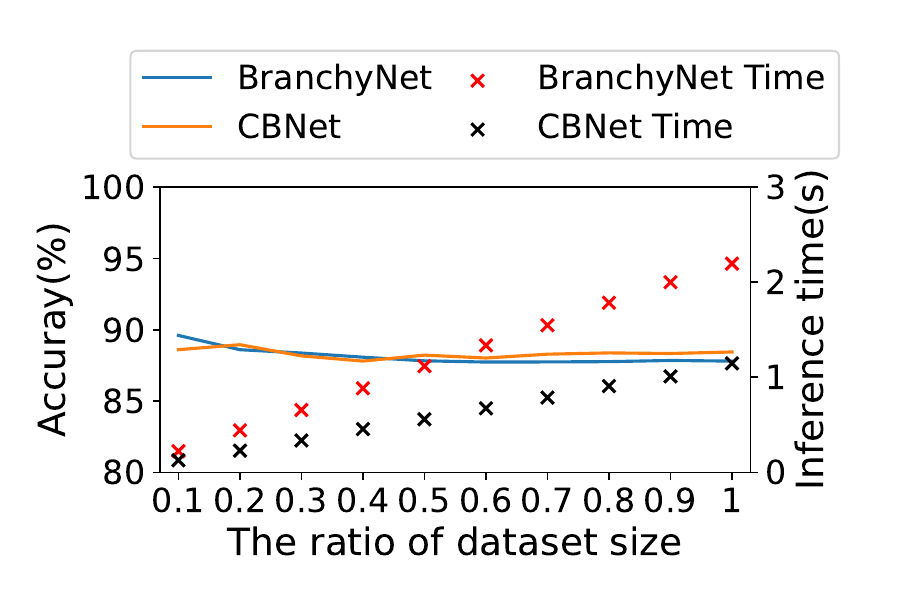}
    \caption{Google cloud instance with GPU}
    \label{fig:scale_GPU_fmnist}
    \end{subfigure}
    \caption{Scalability analysis with FMNIST dataset on each hardware platform.}
    \label{fig:scale_fmnist}
\end{figure*}

\begin{figure*}[htb]
    \begin{subfigure}[b]{0.33\textwidth}
      \centering
    \includegraphics[width=\textwidth]{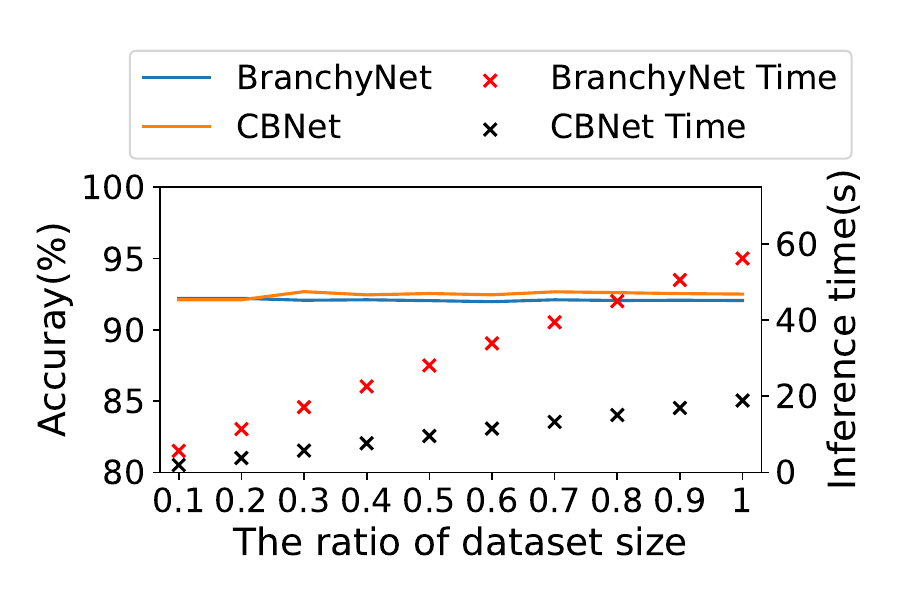}
    \caption{Raspberry Pi}
    \label{fig:Raspberrpi_scale_kmnist}
    \end{subfigure}
    \begin{subfigure}[b]{0.33\textwidth}
      \centering
    \includegraphics[width=\textwidth]{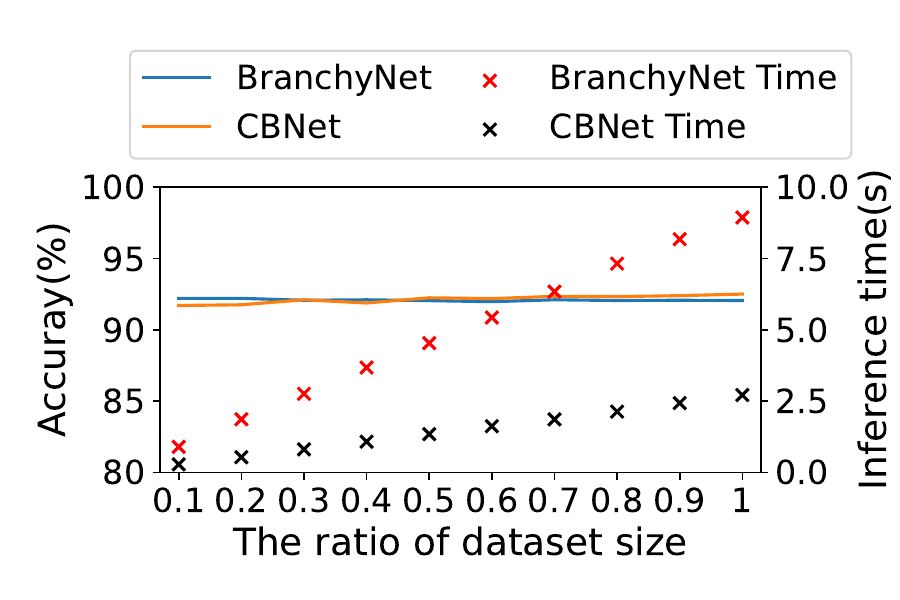}
    \caption{Google cloud instance without GPU}
    \label{fig:local_scale_kmnist}
    \end{subfigure}
    \begin{subfigure}[b]{0.33\textwidth}
      \centering
    \includegraphics[width=\textwidth]{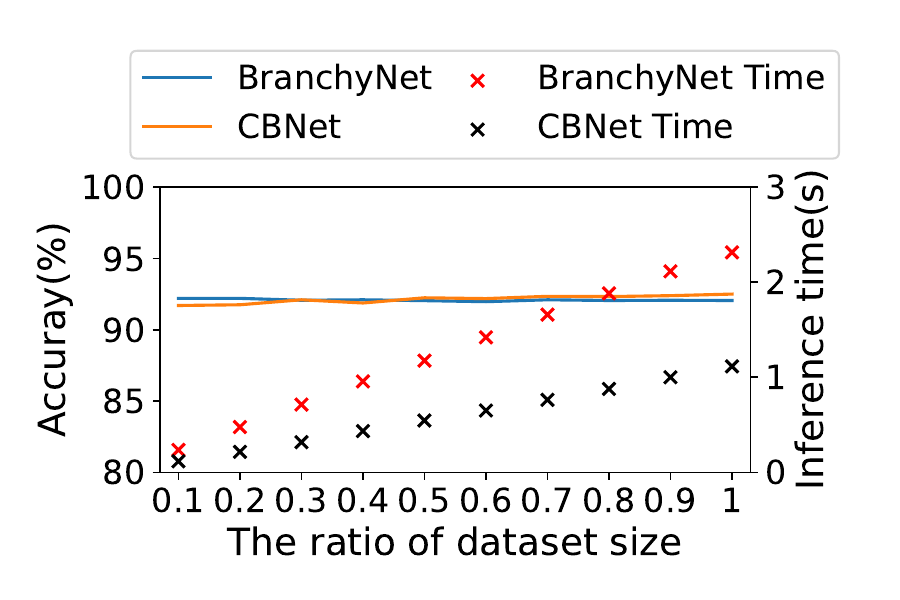}
    \caption{Google cloud instance with GPU}
    \label{fig:scale_GPU_kmnist}
    \end{subfigure}
    \caption{Scalability analysis with KMNIST dataset on each hardware platform.}
    \label{fig:scale_kmnist}
\end{figure*}

\subsection{Energy Efficiency}

Table~\ref{tab:CBNet_Performance} compares the energy savings achieved by CBNet and BranchyNet over the baseline LeNet model on various datasets and devices. The results reported are an average of three runs of each experiment. CBNet achieves 80\% to 85\% savings in energy usage compared to LeNet. It uses 17\% to 79\% less energy than BranchyNet. The energy savings against LeNet in the cases of the MNIST datasets is in the range of 17\%  to 85\%. Whereas in the cases of FMNIST and KMNIST datasets, CBNet's energy savings is in the range of 65\% to 85\%. The difference in the worst-case energy savings is due to the fact that the percentage of hard images present in the FMNIST and KMNIST datasets is higher than in the MNIST datasets. Having more hard images in the dataset provides more opportunities to improve energy efficiency over LeNet and BranchyNet.

%In Figure~\ref{fig:Inference_Energy}, we observe that Truncated BranchyNet has the lowest energy usage and inference latency compared to other approaches. However, this comes with a significant loss in classification accuracy. On the other hand, Ours is able to save energy and reduce inference latency while achieving similar or better classification accuracy than the competing models. This is mainly due to the fact that Ours successfully transforms hard images into easy images so that all images can have quick inference without sacrificing accuracy. 

Notably, the improvement in the inference latency is the dominant factor for reducing energy usage in the cases of Rasberry Pi 4 and GCI without GPU. This is because energy usage is a product of power consumption and inference latency, and there is a negligible difference in the CPU power consumption between various models. However in the case of GCI with GPU, both the improvement in total power consumption and inference latency contribute to the overall energy saving. For each model and dataset, the average CPU power consumption is 17.7 Watts while the average GPU power consumption is six times higher (79 Watts). 
%In the case of GCI with GPU, CBNet uses the CPU 65\% of the time running the \emph{converting autoencoder}, and uses the GPU 35\% of the time for DNN inference. For example, it uses the CPU for 0.68 seconds and the GPU for 0.36 seconds to process 10000 test samples in the FMNIST dataset. As a result, it consumes 39.86 (0.69 seconds $\times$ 17.7 Watts + 0.35 seconds $\times$ 79 Watts) Joules of energy. Whereas BranchyNet uses the GPU for 2.25 seconds to process the same dataset, thus consuming 177.75 (2.25 seconds $\times$ 79 Watts) Joules of energy. 

%energy when using KMNIST datasets and improve 2\% accuracy. For MNIST and QMNIST, CBNet can improve 50\% energy efficiency and only have 0.5\% accuracy loss compared with BranchyNet. For FMNIST and KMNIST, CBNet can save 60\% to 75\% additional energy and improve at least 2\% accuracy compared with BranchyNet. CBNet can improve 0.5\% to 5\% accuracy and bring only 5\% additional energy consumption compared with Truncated BranchyNet. 

%Fig. \ref{fig:Inference_time} shows the inference time for each model. The CBNet compared with BranchyNet has a better inference time. The CBNet has a higher inference time with Truncated BranchyNet because CBNet uses an autoencoder. 

\subsection{Scalability Analysis}

%Fig. \ref{fig:scale_mnist}, Fig. \ref{fig:scale_fmnist}, Fig. \ref{fig:scale_kmnist}  and Fig. \ref{fig:scale_q_mnist} show the accuracy and inference time when data set is gradually incremented with a fixed ratio. CBNet has a better performance than BranchyNet when running in the Raspberry Pi or Google cloud server without GPU. For running in the Google cloud server with GPU, our autoencoder's program is not optimized for running with GPU. Therefore, the improvement for inference time is very small. We will explore it in the future. 

We analyzed the scalability of CBNet by evaluating the impact of increasing dataset size on total inference time and classification accuracy as shown in Figures~\ref{fig:scale_mnist} to \ref{fig:scale_kmnist}. The data size ratio indicates the fraction of the total dataset used in the experiment. For example, a ratio of 0.1 means that only 10\% of the test images available in the dataset were used. We ensured that the proportion of hard test images used in each experiment remained roughly the same. Each experiment was repeated three times, and the average result is reported. We observe that the difference in the inference latency between CBNet and BranchyNet increases with the size of the dataset. This trend is more prominent in the cases of FMNIST and KMNIST datasets since these datasets contain more hard images than others, thus providing more opportunities for CBNet to improve the inference latency. The difference in the accuracy between CBNet and BranchyNet does not change much with increasing dataset size.

%% file: conclusion.tex
\section{Conclusion and Future Work} \label{Conclusion}

 %In this paper, we present CB-LeNet, an energy-efficient framework for DNN inferencing at the edge. The main component of CB-LeNet is an autoencoder that transforms hard images into easy images. These transformed images are then fed to an early-exiting  DNN  for faster inferencing. CB-LeNet ensures that all inputs are  inferenced  efficiently  and  accurately  by  taking  early  exits in  a  DNN.  We have conducted our experiments using four different daatsets. We have Experimental  results  using  four  image-classificationdatasets  on  a  Rasberry  Pi  4  and  Google  cloud  server  show  that CB-LeNet  reduces  the  total  energy  usage  associated  with  DNNinferencing by 40\% and 72\% respectively compared to an existingearly-exit  DNN  (BranchyNet)  without  sacrificing  accuracy. In the future, we will extend this work to include more complex datasets and optimize autoencoder to support GPU.
 
With the increasing trend of deploying deep learning models at the edge, recent research has focused on reducing the latency and energy usage of DNN inference on resource-constrained edge devices. One promising line of research deals with early-exit DNNs where only the result from computing the initial layers is used to classify less complex (easy) images. However, this approach can be inefficient when there is a large number of hard images in the dataset. In this paper, we have presented CBNet, an efficient framework for DNN inference at the Edge. It features a ``converting" autoencoder that transforms hard images into easy images, which are then fed to a lightweight DNN for inference. We have developed a general approach to training the autoencoder and obtaining the corresponding lightweight DNN. We have conducted our experiments using three popular image classification datasets and heterogeneous hardware platforms. Experiments conducted on a Raspberry Pi 4, a Google Cloud instance, and an instance with Nvidia Tesla K80 GPU show that CBNet achieves significant energy savings and speedup in inference latency compared to competing techniques, while maintaining similar accuracy. Our ongoing work shows promising initial results in extending the applicability of converting autoencoders to non-early-exiting DNNs. In the future, we aim to expand this endeavor by incorporating more complex datasets and DNN architectures such as AlexNet\cite{AlexNet} and ResNet\cite{ResNet}. Our future goal is also to generalize our approach, eliminating the dependency on branchynet for easy-hard classification while removing the decoder block. We also plan to utilize physical Raspberry Pi 4 devices in conjunction with an energy meter\cite{UM34C} capable of accurately measuring energy consumption in real time.

%In the future, we aim to expand this endeavor by incorporating more complex datasets and DNN architectures such as AlexNet\cite{AlexNet} and ResNet\cite{ResNet}. Our future goal is also to generalize our approach, eliminating the dependency on branchynet for simplified hard classification while removing the decoder block.

%In the future, we intend to extend this work to include more complex datasets and DNN architectures such as AlexNet\cite{AlexNet} and ResNet\cite{ResNet}.